\documentclass[10pt]{article} 
\usepackage[preprint]{tmlr}

\usepackage{amsmath,amsfonts,bm}

\def\eqref#1{equation~\ref{#1}}

\def\1{\bm{1}}

\DeclareMathAlphabet{\mathsfit}{\encodingdefault}{\sfdefault}{m}{sl}
\SetMathAlphabet{\mathsfit}{bold}{\encodingdefault}{\sfdefault}{bx}{n}

\definecolor{mydarkblue}{rgb}{0,0.08,0.45}
\usepackage[colorlinks,citecolor=mydarkblue,urlcolor=mydarkblue,linkcolor=mydarkblue]{hyperref}
\usepackage{url}
\usepackage{graphicx}
\usepackage{enumitem}
\usepackage{subfig}

\title{Explainability for Large Language Models: A Survey}

\author{\vspace{1.3mm}\name Haiyan Zhao\hspace{0.5mm}$^{1}$ \email hz54@njit.edu \\
    \vspace{1.3mm}\name Hanjie Chen\hspace{0.5mm}$^{2, 8}$ \email  hanjie@rice.edu \\
    \vspace{1.3mm}\name Fan Yang\hspace{0.5mm}$^{3}$ \email  yangfan@wfu.edu \\
    \vspace{1.3mm}\name Ninghao Liu\hspace{0.5mm}$^{4}$ \email  ninghao.liu@uga.edu \\
    \vspace{1.3mm}\name Huiqi Deng\hspace{0.5mm}$^{5}$ \email denghq7@sjtu.edu.cn \\
    \vspace{1.3mm}\name Hengyi Cai\hspace{0.5mm}$^{6}$ \email hengyi1995@gmail.com \\
    \vspace{1.3mm}\name Shuaiqiang Wang \hspace{0.5mm}$^{7}$ \email  wangshuaiqiang@baidu.com \\
    \vspace{1.3mm}\name Dawei Yin \hspace{0.5mm}$^{7}$ \email yindawei@acm.org \\
    \vspace{3mm}\name Mengnan Du\hspace{0.5mm}$^{1}$ \email mengnan.du@njit.edu \\
    \addr
    $^{1}$New Jersey Institute of Technology \hspace{1mm}
    $^{2}$Johns Hopkins University \hspace{1mm}
    $^{3}$Wake Forest University \hspace{1mm} 
    $^{4}$University of Georgia \hspace{1mm} 
    $^{5}$Shanghai Jiao Tong University \hspace{1mm} 
    $^{6}$Institute of Computing Technology, CAS \hspace{1mm} 
    $^{7}$Baidu Inc. \hspace{1mm} 
    $^{8}$Rice University
}

\def\month{MM}  
\def\year{YYYY} 
\def\openreview{\url{https://openreview.net/forum?id=XXXX}} 

\begin{document}
\maketitle
\begin{abstract}
Large language models (LLMs) have demonstrated impressive capabilities in natural language processing. However, their internal mechanisms are still unclear and this lack of transparency poses unwanted risks for downstream applications. Therefore, understanding and explaining these models is crucial for elucidating their behaviors, limitations, and social impacts. In this paper, we introduce a taxonomy of explainability techniques and provide a structured overview of methods for explaining Transformer-based language models. We categorize techniques based on the training paradigms of LLMs: traditional fine-tuning-based paradigm and prompting-based paradigm. For each paradigm, we summarize the goals and dominant approaches for generating local explanations of individual predictions and global explanations of overall model knowledge. We also discuss metrics for evaluating generated explanations, and discuss how explanations can be leveraged to debug models and improve performance. Lastly, we examine key challenges and emerging opportunities for explanation techniques in the era of LLMs in comparison to conventional deep learning models.

\end{abstract}

\section{Introduction}
\label{sec:introduction}
Large language models (LLMs), such as BERT~\citep{devlin2018bert}, GPT-3~\citep{brown2020language}, GPT-4~\citep{openai2023gpt4}, LLaMA-2~\citep{touvron2023llama2}, and Claude~\citep{AnthropicAI2023}, have demonstrated impressive performance across a wide range of natural language processing (NLP) tasks.
Major technology companies, such as Microsoft, Google, and Baidu, have deployed LLMs in their commercial products and services to enhance functionality. For instance, Microsoft leverages GPT-3.5 to improve search relevance ranking in new Bing~\citep{microsoftReinventingSearch}.
Since LLMs are notoriously complex ``black-box'' systems, their inner working mechanisms are opaque, and the high complexity makes model interpretation much challenging. This lack of model transparency can lead to the generation of harmful content or hallucinations in some cases~\citep{weidinger2021ethical}. 
Therefore, it is critical to develop explainability to shed light on how these powerful models work.

Explainability\footnote{In the following sections, we use explainability and interpretability interchangeably.} refers to the ability to explain or present the behavior of models in human-understandable terms~\citep{doshi2017towards,du2019techniques}. Improving the explainability of LLMs is crucial for two key reasons. First, for general end users, explainability builds appropriate trust by elucidating the reasoning mechanism behind model predictions in an understandable manner, without requiring technical expertise. 
With that, end users are able to understand the capabilities, limitations, and potential flaws of LLMs. Second, for researchers and developers, explaining model behaviors provides insight to identify unintended biases, risks, and areas for performance improvements. 
In other words, explainability acts as a debugging aid to quickly advance model performance on downstream tasks~\citep{strobelt_seq2seq-vis_2018, bastings_will_2022, yuksekgonul_post-hoc_2023}. 
It facilitates the ability to track model capabilities over time, make comparisons between different models, and develop reliable, ethical, and safe models for real-world deployment.

\begin{figure}[t]
  \centering
  \includegraphics[width=0.9\textwidth]{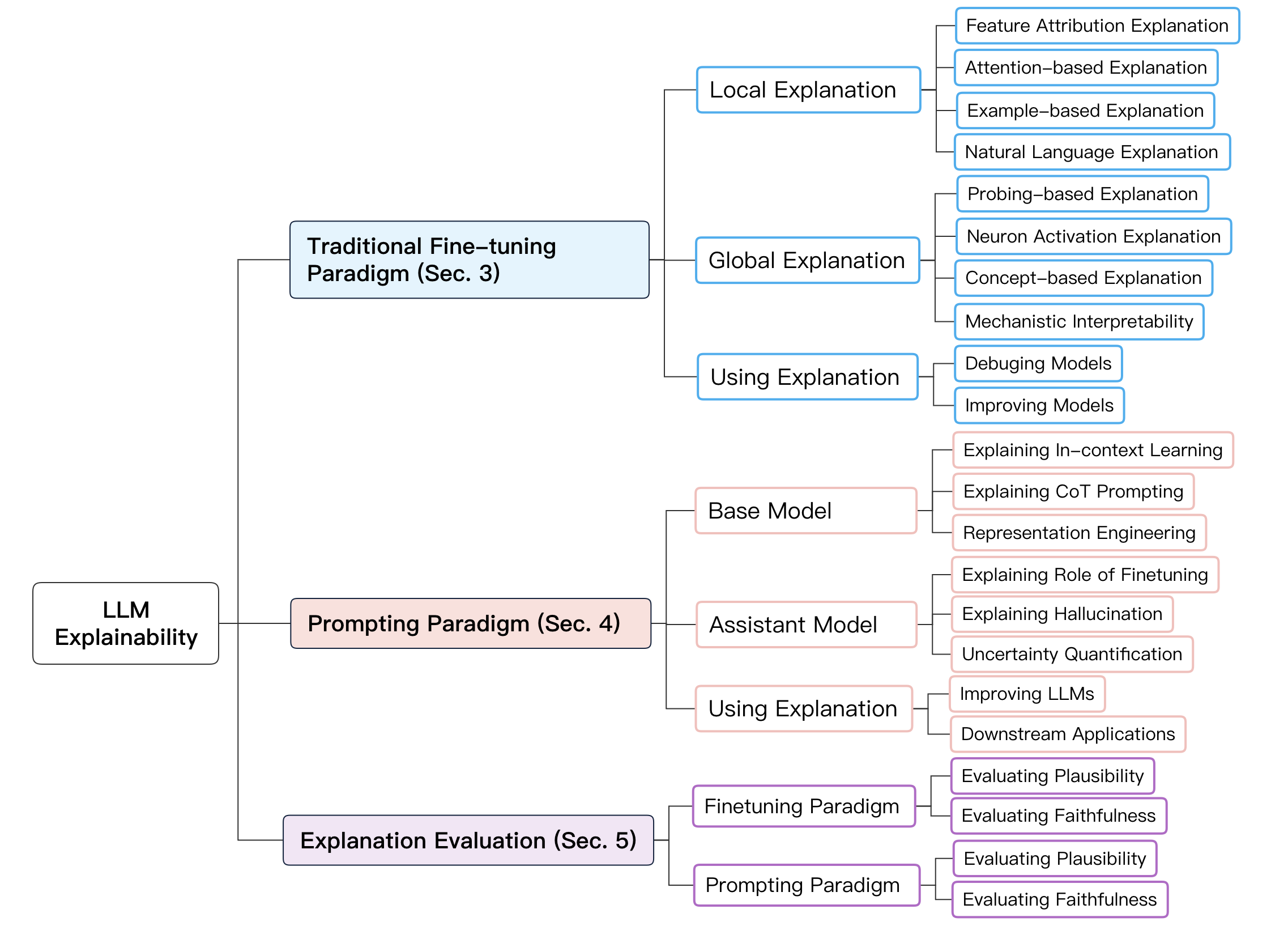}
  \vspace{-5pt}
  \caption{We categorize LLM explainability into two major paradigms. Based on this categorization, we summarize different kinds of explainability techniques associated with LLMs belonging to these two paradigms. 
  We also discuss evaluations for the generated explanations under the two paradigms.
  }
  \label{fig:llm_survey_framework}
  \vspace{-5pt}
\end{figure}

{In contrast to traditional deep learning models, the scale of LLMs in terms of parameters and training data introduces both complex challenges and exciting opportunities for explainability research.
Firstly, as models become larger, understanding and interpreting their decision-making processes grows more difficult due to increased internal complexity and the vastness of training data. This complexity also necessitates significant computational resources to generate explanations. On the one hand, traditionally practical feature attribution techniques, such as gradient-based methods~\citep{sundararajan2017axiomatic} and SHAP values~\citep{lundberg2017unified}, could demand substantial computational power to explain LLMs with billions of parameters. This makes these explanation techniques less practical for real-world applications that end-users can utilize. On the other hand, this increased complexity makes in-depth analysis challenging, obstructing debugging and diagnosing of the models. Furthermore, comprehending the unique capabilities of LLMs in in-context learning~\citep{li2023understanding} and chain-of-thought prompting~\citep{wu2023analyzing}, as well as the phenomenon of hallucination, is indispensable to explain and improve models. Secondly, this scaling also spurs innovation in interpretability techniques and offers richer insights into model behavior. For instance, LLMs could provide chain-of-thought explanations for their own decision-making processes. Additionally, recent research finds LLMs can serve as tools to provide post-hoc explanations for predictions made by other machine learning models~\citep{kroeger2023large}. To better understand and enhance LLMs, it is imperative to review available explainability techniques and develop an understanding of potential future directions.}

In this paper, we provide a comprehensive overview of methods for interpreting Transformer-based language models. In Section~\ref{sec:train-parad-llms}, we introduce the two main paradigms in applying LLMs: 1) the traditional downstream fine-tuning paradigm and 2) the prompting paradigm. Based on this categorization, we review explainability methods for fine-tuned LLMs in Section~\ref{sec:pre-train-downstr}, and prompted LLMs in Section~\ref{sec:prompting-paradigm}. In Section~\ref{sec:eval-expl}, we discuss the evaluation of explainability methods. Finally, in Section~\ref{sec:research-challenges}, we further discuss research challenges in explaining LLMs compared to traditional {deep} learning models and provide insight on potential future research directions. This paper aims to comprehensively organize recent research progress on interpreting complex language models. 

\section{Training Paradigms of LLMs}
\label{sec:train-parad-llms}
The training of LLMs can be broadly categorized into two paradigms, \textit{traditional fine-tuning} and \textit{prompting}, based on how they are used for adapting to downstream tasks. Due to the substantial distinctions between the two paradigms, various types of explanations have been proposed respectively (shown in Figure~\ref{fig:llm_survey_framework}).

\subsection{Traditional Fine-Tuning Paradigm}
In this paradigm, a language model is first pre-trained on a large corpus of unlabeled text data, and then fine-tuned on a set of labeled data from a specific downstream domain, such as SST-2, MNLI, and QQP on the GLUE benchmark~\citep{wang2018glue}. During fine-tuning, it is easy to add fully connected layers above the final encoder layer of the language model, allowing it to adapt to various downstream tasks~\citep{rogers2021primer}. 
This paradigm has shown success for medium-sized language models, typically containing up to one billion parameters. Examples include BERT~\citep{devlin2018bert}, RoBERTa~\citep{liu2019roberta}, ELECTRA~\citep{clark2020electra}, DeBERTa~\citep{hedeberta}, etc. Explanations on this paradigm focus on two key areas: 1) Understanding how the self-supervised pre-training enables models to 
acquire a foundational understanding of language (e.g., syntax, semantics, and contextual relationships); and 2) Analyzing how the fine-tuning process 
equips these pre-trained models with the capability to effectively solve downstream tasks.

\subsection{Prompting Paradigm}
The prompting paradigm involves using prompts, such as natural language sentences with blanks for the model to fill in, to enable zero-shot or few-shot learning without requiring additional training data. Models under this paradigm can be categorized into two types, based on their development stages:

\noindent\textbf{Base Model:} As LLMs scale up in size and training data, they exhibit impressive new capabilities without requiring additional training data. One such capability is few-shot learning through prompting.
This type of paradigm usually works on huge-size language models (with billions of parameters) such as GPT-3~\citep{brown2020language}, OPT~\citep{zhang2022opt}, LLaMA-1~\citep{touvron2023llama}, LLaMA-2~\citep{touvron2023llama2}, Falcon~\citep{almazrouei2023falcon}. These models are called \emph{foundation models or base models}\footnote{In this work, we refer foundation models to very large scale models such as LLaMA-2.}, which can chat with users without further alignment with human preferences. Large-scale models typically fit into this paradigm, with size exceeding 1B. For example, LLaMA-2~\citep{touvron2023llama2} has up to 70B parameters. Explanations for base models aim to understand how models learn to leverage its pre-trained knowledge in response to the prompts.

\begin{figure}[t]
  \centering
  \includegraphics[width=0.85\linewidth]{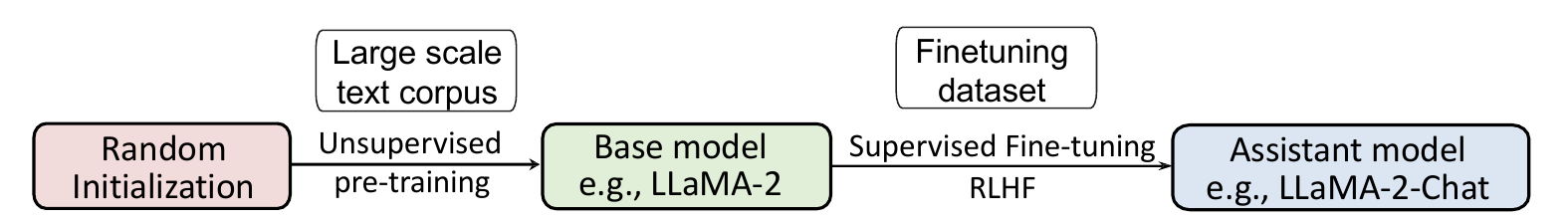}
  \caption{LLMs undergo unsupervised pre-training with random initialization to create a base model. The base model can then be fine-tuned through instruction tuning and RLHF to produce the assistant model.}
  \label{fig:chat-model}
\end{figure}

\noindent\textbf{Assistant Model:} There are two major limitations of the base models: 1) they cannot follow user instructions as the pre-training data contains few instruction-response examples, and 2) they tend to generate biased and toxic content~\citep{carlini2023aligned}.
To address these limitations, the base models are further fine-tuned with supervised fine-tuning (see Figure~\ref{fig:chat-model}) to achieve human-level abilities, such as open-domain dialogue. The key idea is to align the model's responses with human feedback and preferences. The most typical way for this process is through instruction tuning via (prompts, response) demonstration pairs and Reinforcement Learning from Human Feedback (RLHF). Models are trained with natural language feedback to carry out complex, multi-turn conversations. 
Models belonging to this family include GPT-3.5 and GPT-4~\citep{bubeck2023sparks} by OpenAI, Claude by Anthropic~\citep{AnthropicAI2023}, and open-source models such as LLaMA-2-Chat by Meta~\citep{touvron2023llama2}, Alpaca~\citep{taori2023alpaca} and Vicuna~\citep{vicuna2023}. 
These models can be called \emph{assistant models, chat assistants, or dialogue models}.
Explanations here focus on understanding how models learn open-ended interactive behaviors from conversations.

\section{Explanation for Traditional Fine-Tuning Paradigm }
\label{sec:pre-train-downstr}
In this section, we review explanation techniques for LLMs trained with the pre-training and downstream fine-tuning paradigms. First, we introduce approaches to provide local explanations (Section~\ref{sec:local-explanation}) and global explanations (Section~\ref{sec:global-explanation}). Here, \textit{local explanation} aims to provide an understanding of how a language model makes a prediction for a specific input instance, while \textit{global explanation} aims to provide a broad understanding of how the LLM works overall. Next, we discuss how explanations can be used to debug and improve models (Section~\ref{sec:using-expl-impr}).

\subsection{Local Explanation}
\label{sec:local-explanation}
\begin{figure}[t]
    \centering
    \includegraphics[width = 0.98\textwidth]{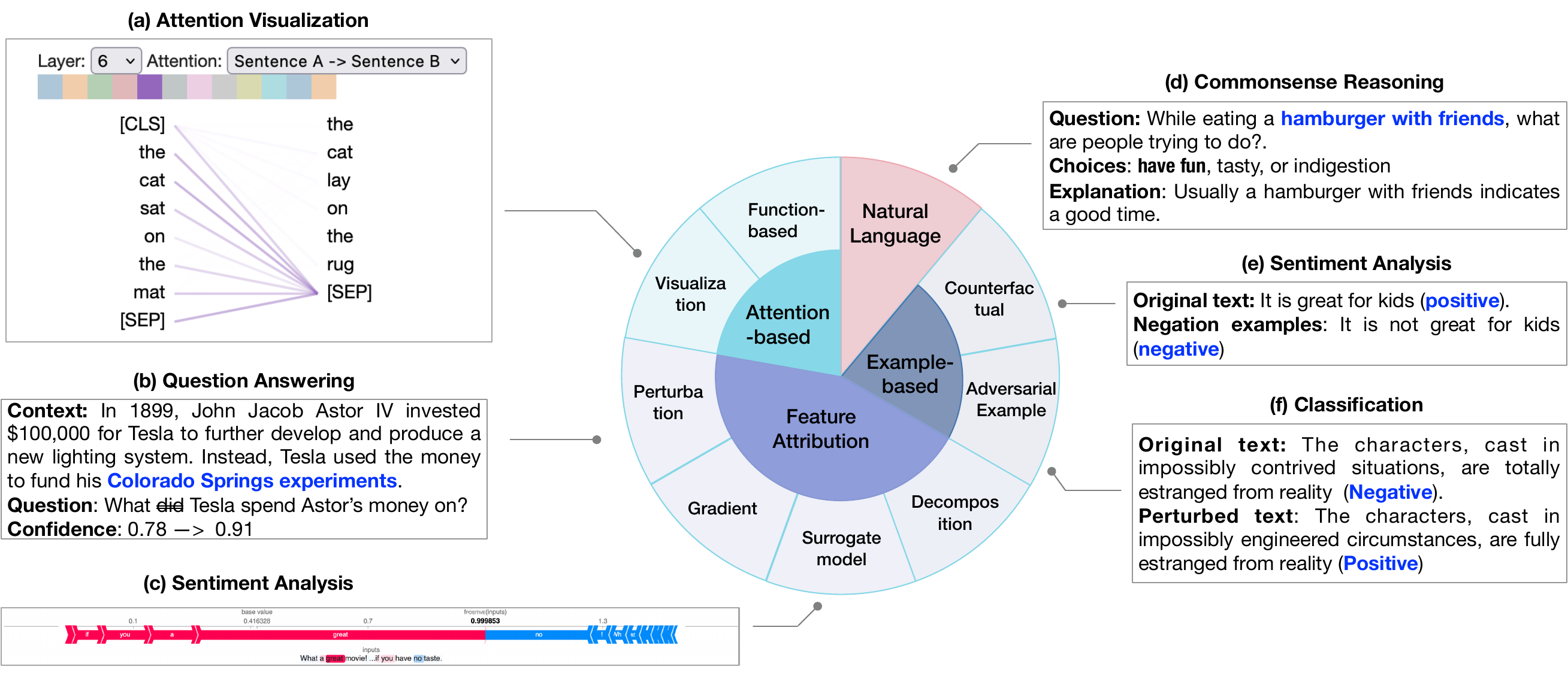}
    \caption{Local explanation is composed of four subareas. The organization of each subarea and examples for certain individual explanation methodology have been given. (a) Bipartite graph attention representation for attention matrix between sentence A and sentence B at the 6th layer~\citep{vig_bertviz_2019}; (b) Perturb the question by deleting ``did'', the confidence of the answer ``Colorado Springs experiments'' has even increased for the reduced question while the answer is nonsense for human~\citep{feng_pathologies_2018}; (c) Shapley values for transformer-based language models~\citep{chen_algorithms_2023}; (d) Provide explanation to the important components of input text to assist in commonsense reasoning~\citep{rajani_explain_2019}; (e) Provide negative examples of input text to test model's ability in sentiment prediction and can also be used to improve model performance~\citep{wu_polyjuice_2021}; (f) Change the input text in an imperceptible way for humans but the classification is distracted from the original~\citep{jin2019bert}.}
    \label{fig:local}
\end{figure}

The first category of explanations refers to explaining the predictions generated by LLM. Let us consider a scenario where we have a language model and we input a specific text into the model. The model then produces a classification output, such as sentiment classification or a prediction for the next token. In this scenario, the role of explanation is to clarify the process by which the model generated the particular classification or token prediction. Since the goal is to explain how the LLM makes the prediction for a specific input, we call it the \textit{local explanation}. This category encompasses four main streams of approaches for generating explanations including feature attribution-based explanation, attention-based explanation, example-based explanation, and natural language explanation (see Figure~\ref{fig:local}).

\subsubsection{Feature Attribution-Based Explanation}
\label{sec:feat-attr-based}
Feature attribution methods aim to measure the relevance of each input feature (e.g., words, phrases, text spans) to a model's prediction. Given an input text $\boldsymbol{x}$ comprised of $n$ word features $\{x_1, x_2, ..., x_n\}$, a fine-tuned language model $f$ generates an output $f(\boldsymbol{x})$. Attribution methods assign a relevance score $R(x_i)$ to the input word feature $x_i$ to reflect its contribution to the model prediction $f(\boldsymbol{x})$. The methods that follow this strategy can be mainly categorized into four types: perturbation-based methods, gradient-based methods, surrogate models, and decomposition-based methods.

\paragraph{Perturbation-Based Explanation}
Perturbation-based methods work by perturbing input examples such as removing, masking, or altering input features and evaluating model output changes. The most straightforward strategy is \textit{leave-one-out}, which perturbs inputs by removing features at various levels including embedding vectors, hidden units~\citep{li_understanding_2017}, words~\citep{li_visualizing_2016}, tokens and spans~\citep{wu_perturbed_2020} to measure feature importance. The basic idea is to remove the minimum set of inputs to change the model prediction. The set of inputs is selected with a variety of metrics such as confidence score or reinforcement learning. However, this removal strategy assumes that input features are independent and ignores correlations among them. Additionally, methods based on the confidence score can fail due to pathological behaviors of overconfident models~\citep{feng_pathologies_2018}.
For example, models can maintain high-confidence predictions even when the reduced inputs are nonsensical.
This overconfidence issue can be mitigated via regularization with regular examples, label smoothing, and fine-tuning models' confidence~\citep{feng_pathologies_2018}. Besides, current perturbation methods tend to generate out-of-distribution data. This can be alleviated by constraining the perturbed data to remain close to the original data distribution~\citep{qiu_resisting_2021}.

\paragraph{Gradient-Based Explanation}
Gradient-based attribution techniques determine the importance of each input feature by analyzing the partial derivatives of the output with respect to each input dimension. The magnitude of derivatives reflects the sensitivity of the output to changes in the input. The basic formulation of raw gradient methods is described as 
$
  \boldsymbol{s}_j=\frac{\partial f(\boldsymbol{x})}{\partial \boldsymbol{x}_j},
$
where $f(\boldsymbol{x})$ is the prediction function of the network and $\boldsymbol{x}_j$ denotes the input vector. This scheme has also been improved as gradient $\times$ input~\citep{kindermans_reliability_2017} and has been used in various explanation tasks, such as computing the token-level attribution score~\citep{mohebbi_exploring_2021}. However, vanilla gradient-based methods have some major limitations. First, they do not satisfy the input invariance, meaning that input transformations such as constant shift can generate misleading attributions without affecting the model prediction~\citep{kindermans_reliability_2017}. Second, they fail to deal with zero-valued inputs. Third, they suffer from gradient saturation where large gradients dominate and obscure smaller gradients. The difference-from-reference approaches, such as integrated gradients (IG), are believed to be a good fit to solve these challenges by satisfying more axioms for attributions~\citep{sundararajan2017axiomatic}. The fundamental mechanism of IG and its variants is to accumulate the gradients obtained as the input is interpolated between a reference point and the actual input. The baseline reference point is critical for reliable evaluation, but the criteria for choosing an appropriate baseline remain unclear. Some use noise or synthetic reference with training data, but performance cannot be guaranteed~\citep{lundstrom_rigorous_2022}. In addition, IG struggles to capture output changes in saturated regions and should focus on unsaturated regions~\citep{miglani_investigating_2020}. Another challenge of IG is the computational overhead to achieve high-quality integrals. Since IG integrates along a straight line path that does not fit well the discrete word embedding space, variants have been developed to adapt it for language models~\citep{sikdar_integrated_2021, sanyal_discretized_2021, enguehard_sequential_2023}.

\paragraph{Surrogate Models}
Surrogate models methods use simpler, more human-comprehensible models to explain individual predictions of black-box models. These surrogate models include decision trees, linear models, decision rules, and other white-box models that are inherently more understandable to humans. The explanation models need to satisfy additivity, meaning that the total impact of the prediction should equal the sum of the individual impacts of each explanatory factor. Also, the choice of interpretable representations matters. Unlike raw features, these representations should be powerful enough to generate explanations yet still understandable and meaningful to human beings. An early representative local explanation method called LIME~\citep{ribeiro_why_2016} employs this paradigm. To generate explanations for a specific instance, the surrogate model is trained on data sampled locally around that instance to approximate the behavior of the original complex model in the local region. However, it is shown that LIME does not satisfy some properties of additive attribution, such as local accuracy, consistency, and missingness~\citep{lundberg_unified_2017}.
SHAP is another framework that satisfies the desirable properties of additive attribution methods~\citep{lundberg_unified_2017}. It treats features as players in a cooperative prediction game and assigns each subset of features a value reflecting their contribution to the model prediction. Instead of building a local explanation model per instance, SHAP computes Shapley values~\citep{shapley1953value} using the entire dataset. Challenges in applying SHAP include choosing appropriate methods for removing features and efficiently estimating Shapley values. Feature removal can be done by replacing values with baselines like zeros, means, or samples from a distribution, but it is unclear how to pick the right baseline. Estimating Shapley values also faces computational complexity exponential in the number of features. Approximation strategies including weighted linear regression, permutation, and other model-specific methods have been adopted~\citep{chen_algorithms_2023} to estimate Shapley values. Despite complexity, SHAP remains popular and widely used due to its expressiveness for large deep models. To adapt SHAP to Transformer-based language models, approaches such as TransSHAP have been proposed~\citep{chen_algorithms_2023,kokalj_bert_2021}. TransSHAP mainly focuses on adapting SHAP to sub-word text input and providing sequential visualization explanations that are well suited for understanding how LLMs make predictions.

\paragraph{Decomposition-Based Methods}
Decomposition techniques aim to break down the relevance score into linear contributions from the input. Some work assign relevance score directly from the final output layer to the input~\citep{du_attribution_2019}. The other line of work attributes relevance score layer by layer from the final output layer toward the input. Layer-wise relevance propagation (LRP)~\citep{montavon2019layer} and Taylor-type decomposition approaches (DTD)~\citep{montavon_explaining_2015} are two classes of commonly used methods. The general idea is to decompose the relevance score $R_j^{(l+1)}$ of neuron $j$ in layer $l+1$ to each of its input neuron $i$ in layer $l$, which can be formulated as:
$
R_j^{(l+1)}=\sum_i R_{i \leftarrow j}^{(l, l+1)}.
$
The key difference is in the relevance propagation rules used by LRP versus DTD. These methods can be applied to break down relevance scores into contributions from model components such as attention heads~\citep{voita_analyzing_2019}, tokens, and neuron activation~\citep{voita_analyzing_2021}. Both methods have been applied to derive the relevance score of inputs in Transformer-based models~\citep{wu_explaining_2021,chefer_transformer_2021}.

\subsubsection{Attention-Based Explanation}
\label{sec:attent-based-expl}
Attention mechanism is often viewed as a way to attend to the most relevant part of inputs. Intuitively, attention may capture meaningful correlations between intermediate states of input that can explain the model's predictions. Many existing approaches try to explain models solely based on the attention weights or by analyzing the knowledge encoded in the attention. These explanation techniques can be categorized into three main groups: visualization methods, function-based methods, and probing-based methods. As probing-based techniques are usually employed to learn global explanations, they are discussed in Section~\ref{sec:probing}. In addition, there is an extensive debate in research on whether attention weights are actually suitable for explanations. This topic will be covered later in the discussion.

\begin{figure}[t]
    \centering
    \subfloat[Bipartite Graph]{\includegraphics[height = 5cm]{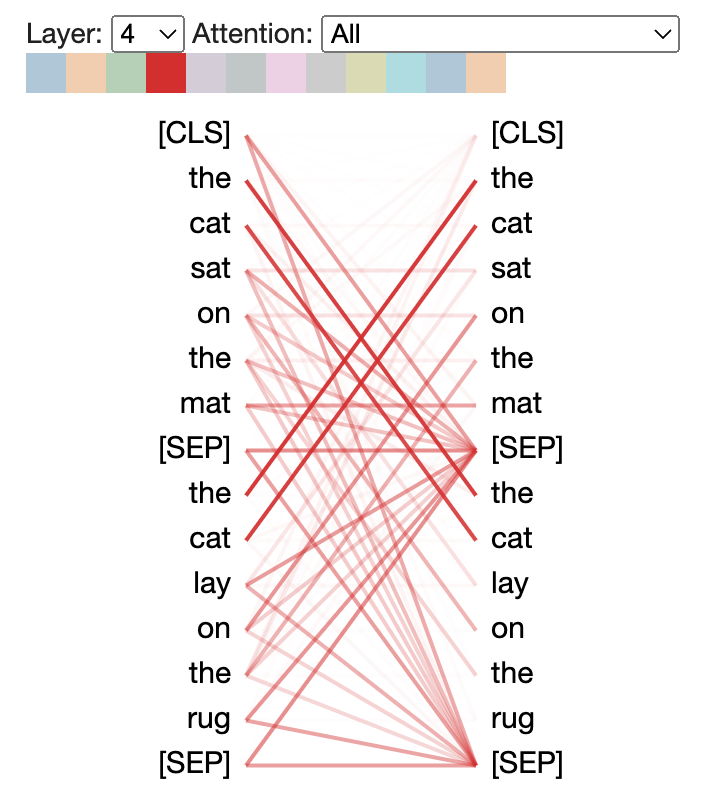}
    }\qquad
    \subfloat[Heatmap]{\includegraphics[height = 5cm]{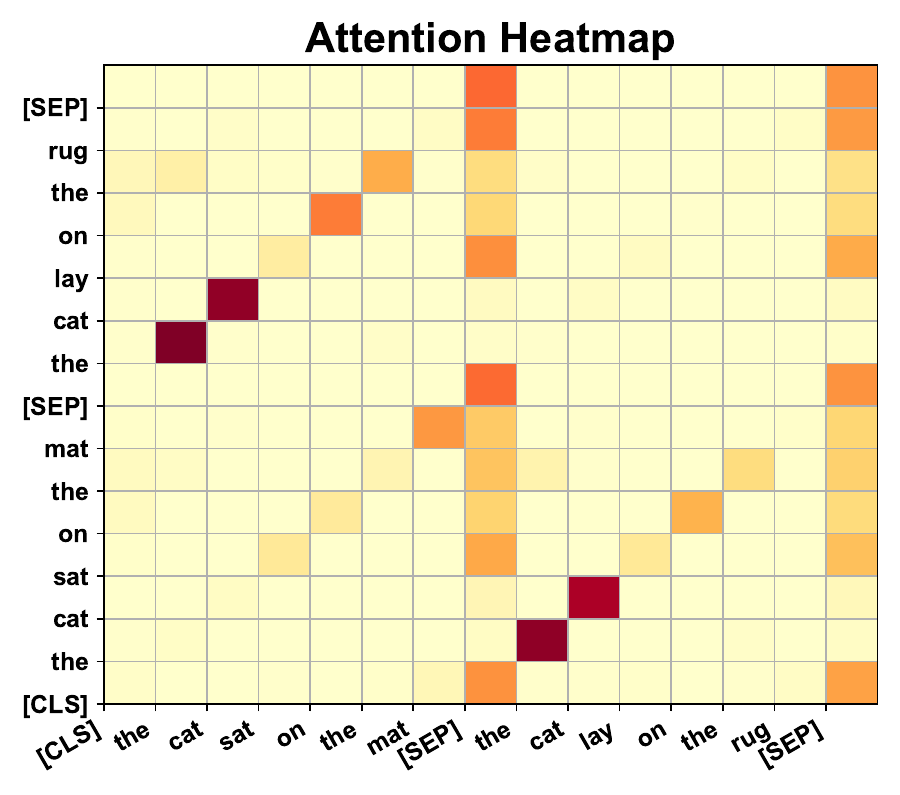}}
    \caption{Bipartite graph attention representation and heatmap for attention matrix.}
    \label{fig:visualization}
\end{figure}

\paragraph{Visualizations}
Visualizing attentions provides an intuitive way to understand how models work by showing attention patterns and statistics. Common techniques involve visualizing attention heads for a single input using bipartite graphs or heatmaps. These two methods are simply disparate visual representation of attentions, one as a graph and the other as a matrix, as illustrated in Figure~\ref{fig:visualization}. Visualization systems differ in their ability to show relationships at multiple scales, by representing attention in various forms for different models. At the input data level, attention scores for each word/token/sentence pairs between the premise sentence and the assumption sentence are shown to evaluate the faithfulness of the model prediction~\citep{vig_bertviz_2019}. Some systems also allow users to manually modify attention weights to observe effects~\citep{jaunet_visqa_2021}. At the neuron level, individual attention heads can be inspected to understand model behaviors~\citep{park_sanvis_2019, vig_bertviz_2019, hoover_exbert_2020, jaunet_visqa_2021}. At the model level, attention across heads and layers is visualized to identify patterns~\citep{park_sanvis_2019, vig_bertviz_2019,yeh_attentionviz_2023}. One notable work focuses on visualizing attention flow to trace the evolution of attention, which can be used to understand information transformation and enable training stage comparison between models~\citep{derose_attention_2020}. Thus, attention visualization provides an explicit, interactive way to diagnose bias, errors, and evaluate decision rules. Interestingly, it also facilitates formulating explanatory hypotheses.

\paragraph{Function-Based methods}
Since raw attention is insufficient to fully explain model predictions, some studies have developed enhanced variants as replacements to identify important attributions for explanation. Gradient is a well-recognized metric for measuring sensitivity and salience, so it is widely incorporated into self-defined attribution scores. These self-designed attribution scores differ in how they define gradients involving attention weights. For example, gradients can be partial derivatives of outputs with respect to attention weights~\citep{barkan_grad-sam_2021} or integrated versions of partial gradients~\citep{hao_self-attention_2021}. The operations between gradients and attention can also vary, such as element-wise products. Overall, these attribution scores that blend attention and gradients generally perform better than using either alone, as they fuse more information that helps to highlight important features and understand networks.

\paragraph{Debate Over Attention}
There is extensive research evaluating attention heads, but the debate over the validity of this approach is unlikely to be resolved soon. The debate stems from several key aspects.
First, some works compare attention-based explanations with those from other methods like LIME. They find that attention often does not identify the most important features for prediction~\citep{serrano_is_2019, jain_attention_2019}. They provide inferior explanations compared to these alternatives~\citep{thorne2019generating} or cannot be correlated with other explanation methods~\citep{jain_attention_2019, liu2020interpretations, ethayarajh_attention_2021}. Second, some directly criticize the usefulness of the attention mechanism in model predictions. They argue that raw attention fails to capture syntactic structures in text and may not contribute to predictions as commonly assumed~\citep{mohankumar_towards_2020}. In addition, raw attention contains redundant information that reduces its reliability in explanations~\citep{bai2021attentions,brunner2019identifiability}. However, other studies contradict these claims. For example, evaluating explanation models for consistency can pose challenges across various approaches, not limited to attention alone~\citep{neely2021order}. 
Besides, manipulation of attention weights without retraining can bias evaluations~\citep{wiegreffe_attention_2019}. Furthermore, attention heads in BERT have been shown to encode syntax effectively~\citep{clark_what_2019}. To make attention explainable, technical solutions have also been explored by optimizing input representation~\citep{mohankumar_towards_2020}, regularizing learning objectives~\citep{moradi2021measuring}, avoiding biased learning~\citep{bai2021attentions} and even incorporating human rationales~\citep{arous2021marta}. But the core reason for the ongoing debates is the lack of well-established evaluation criteria, which will be further discussed in Section~\ref{sec:eva_loc_ex}.

\subsubsection{Example-Based Explanations}
\label{sec:example-based-expl}
Example-based explanations intend to explain model behavior from the perspective of individual instances~\citep{koh2017understanding}. Unlike model-based or feature-based explanations, example-based explanations illustrate how a model's output changes with different inputs. We focus on adversarial examples, counterfactual explanations, and data influence. Adversarial examples are generally synthesized by manipulating less important components of input data. They reveal cases where the model falters or errs, illuminating its weaknesses. In contrast, counterfactual explanations are generated mostly by changing significant parts of input data, and they are popular in scenarios like algorithmic recourse, as providing remedies to a desirable outcome. 
Unlike manipulating inputs, data influence examines how training data impacts a model's predictions on testing data.

\paragraph{Adversarial Example}
Studies show that neural models are highly susceptible to small changes in the input data. These carefully crafted modifications can alter model decisions while barely being noticeable to humans. Adversarial examples are critical in exposing areas where models fail and are usually added to training data to improve robustness and accuracy. Adversarial examples are initially generated by word-level manipulations such as errors, removal, and insertion, which are obvious upon inspection. More advanced token-level perturbation methods like TextFooler~\citep{jin2019bert} have been advanced, which strategically target important words first based on ranking. A candidate word is then chosen based on word embedding similarity, same part-of-speech, sentence semantic similarity, and prediction shift. However, word embedding is limited in sentence representation compared to contextualized representations, often resulting in incoherent pieces. By focusing on contextualized representations, a range of work adopting the mask-then-infill procedure has achieved state-of-the-art performance~\citep{garg_bae_2020, li_contextualized_2021}. They leverage pre-trained masked language models like BERT for perturbations including replacement, insertion, and merging. Typically, a large corpus is employed to train masked language models, generate contextualized representations and obtain token importance. Then models are frozen and perturbation operations are performed on tokens in a ranked order. For replacement, the generated example replaces the masked token. For injection, the new token is inserted into the left or right of the masked token. For merging, a bigram is masked and replaced with one token. SemAttack~\citep{wang_semattack_2022} proposes a more general and effective framework applicable to various embedding spaces including typo space, knowledge space, and contextualized semantic space. The input tokens are first transformed into an embedding space to generate perturbed embeddings that are iteratively optimized to meet attack goals. The experiment shows that replacing 5\% of words reduces BERT's accuracy from 70.6\% to 2.4\% even with defenses in a white-box setting. SemAttack's outstanding attack performance might because it directly manipulates embeddings. 

\paragraph{Counterfactual Explanation}
Counterfactual explanation is a common form of casual explanation, treating the input as the cause of the prediction under the Granger causality. Given an observed input $\boldsymbol{x}$ and a perturbed $\hat{\boldsymbol{x}}$ with certain features changed, the prediction $\boldsymbol{y}$ would change to $\hat{\boldsymbol{y}}$. Counterfactual explanation reveals what would have happened based on certain observed input changes. They are often generated to meet up certain needs such as algorithmic recourse by selecting specific counterfactuals. Examples can be generated by humans or perturbation techniques like paraphrasing or word replacement. A representative generator, Polyjuice~\citep{wu_polyjuice_2021}, supports multiple permutation types for input sentences, such as deletion, negation and shuffling. It can also perturb tokens based on their importance. Polyjuice then finetunes GPT-2 on specific pairs of original and perturbed sentences tailored to downstream tasks, to provide realistic counterfactuals. It generates more extensive counterfactuals with a median speed of 10 seconds per counterfactual, compared to 2 minutes for previous crowd workers-dependent methods~\citep{kaushik_learning_2020}.
Counterfactual explanation generation has been framed as a two-stage approach involving first masking/selecting important tokens and then infilling/editing those tokens~\citep{treviso2023crest,ross2021explaining}. Specifically, MiCE uses gradient-based attribution to select tokens to mask in the first stage and focuses on optimizing for minimal edits through binary search~\citep{ross2021explaining}. In contrast, CREST leverages rationales from a selective rationalization model and relaxes this hard minimality constraint of MiCE. Instead, CREST uses the sparsity budget of the rationalizer to control closeness~\citep{treviso2023crest}. Experiments show that both methods generate high-quality counterfactuals in terms of validity and fluency.

\paragraph{Data Influence} 
{This family of approaches characterize the influence of individual training sample by measuring how much they affect the loss on test points~\citep{yeh2018representer}. The concept originally came from statistics, where it describes how model parameters are affected after removing a particular data point. By observing patterns of influence, we can deepen our understanding of how models make predictions based on their training data. Since researchers have come to recognize the importance of data, several methods have been developed to analyze models from a data-centric perspective. Firstly, influence functions enable us to approximate the concept by measuring loss changes via gradients and Hessian-vector products without the necessity of retraining the model~\citep{koh2017understanding}. \citet{yeh2018representer} decompose a prediction of a test point into a linear combination of training points, where positive values denote excitatory training points and negative values indicate inhibitory points.
Data Shapley employs Monte Carlo and gradient-based methods to quantify the contribution of data points to the predictor performance, and the higher Shapley value tells the desired data type to improve the predictor~\citep{ghorbani2019data}. Another method uses stochastic gradient descent (SGD) and infers the influence of a training point by analyzing minibatches without that point using the Hessian vector of the model parameters~\citep{hara2019data}. Based on such an approach, TracIn derives the influence of training points using the calculus theorem with checkpoints during the training process~\citep{pruthi2020estimating}. However, the aforementioned methods often come with an expensive computational cost even when applied to a medium-sized model. To address this, two key dimensions can be considered: 1) reducing the search space and 2) decreasing the number of approximated parameters in the Hessian vector. \citet{guo2020fastif} also demonstrates the applicability of the influence function in model debugging. Recently, Anthropic has employed the Eigenvalue-corrected Kronecker-Factored Approximate Curvature (EK-FAC) approximation to scale this method to LLMs with 810 million, 6.4 billion, 22 billion, and 52 billion parameters. The result indicates that as model scale increases, influential sequences are better at capturing the reasoning process for queries, whereas smaller models often provide semantically unrelated pieces of information~\citep{grosse2023studying}.
}

\subsubsection{Natural Language Explanation}
\label{sec:natur-lang-expl}
Natural language explanation in NLP refers to explaining a model's decision-making on an input sequence with generated text. The basic approach for generating natural language explanations involves training a language model using both original textual data and human-annotated explanations. The trained language model can then automatically generate explanations in natural language~\citep{rajani_explain_2019}. As explanations provide additional contextual space, they can improve downstream prediction accuracy and perform as a data augmentation technique~\citep{luo_local_2022,yordanov_few-shot_2022}. Apart from the explain-then-predict approach, predict-then-explain and joint predict-explain methods have also been investigated. The choice of methods depends on the purpose of the task. But the reliability of applying generated explanations still necessitates further investigation. It is worth noting that both the techniques introduced in this section and the CoT explanations covered later in Section~\ref{sec:prompting-paradigm} produce natural language explanations. However, the explanations covered here are typically generated by a separate model, while CoT explanations are produced by the LLMs themselves. 

\subsection{Global Explanation}
\label{sec:global-explanation}
Different from local explanations that aim to explain a model's individual predictions, global explanations offer insights into the inner workings of language models.
Global explanations aim to understand what the individual components (neurons, hidden layers, and larger modules) have encoded and explain the knowledge/linguistic properties learned by the individual components.
We examine three main approaches for global explanations: probing methods that analyze model representations and parameters, neuron activation analysis to determine model responsiveness to input, and concept-based methods.

\subsubsection{Probing-Based Explanations}
\label{sec:probing}
The self-supervised pre-training process leads to models that acquire broad linguistic knowledge from the training data. The probing technique refers to methods used to understand the knowledge that LLMs such as BERT have captured.

\paragraph{Classifier-Based Probing}
The basic idea behind classifier-based probing is to train a shallow classifier on top of the pre-trained or fine-tuned language models such as BERT~\citep{devlin2019bert}, T5~\citep{raffel2020exploring}. To perform probing, the parameters of the pre-trained models are first frozen, and the model generates representations for input words, phrases, or sentences and learns parameters like attention weights. These representations and model parameters are fed into a \textit{probe} classifier, whose task is to identify certain linguistic properties or reasoning abilities acquired by the model. Once the probe is trained, it will be evaluated on a holdout dataset. The labeled data comes from available taggers or gold-annotated datasets. Although each probe classifier is often tailored for a certain task, the scheme for training classifiers to probe different knowledge is consistent. Related studies will be presented according to probed model components, i.e. vector representations and model parameters.

We first examine works that study \emph{vector representations} to measure embedded knowledge. In this category, knowledge means either \textit{syntax knowledge} at a low level or \textit{semantic knowledge} at a high level. Studies demonstrate that lower layers are more predictive of word-level syntax, whereas higher layers are more capable of capturing sentence-level syntax and semantic knowledge~\citep{belinkov_evaluating_2017,peters_dissecting_2018, blevins_deep_2018, jawahar_what_2019}. 

Syntactic labels can be further categorized into word- or sentence-level categories. The word-level syntactic labels provide information about each individual word, such as part-of-speech tags, morphological tags, smallest phrase constituent tags, etc. The sentence-level syntactic labels describe attributes of the whole sentence, such as voice (active or passive), tense (past, present, future), and top-level syntactic sequence. For word-level syntax probing, parse trees are often introduced by dependency parser~\citep{dozat_deep_2017} to help extract dependency relations~\citep{tenney_what_2019}. A structural probe is also developed to identify parse trees in a specific vector space by measuring the syntactic distance between all pairs of words with distance metrics~\citep{hewitt_structural_2019, chen_probing_2021}. This demonstrates that syntactic knowledge is embedded in vector representations and is popular for reconstructing dependency trees for probing tasks. However, concerns emerged about whether probing classifiers learn syntax in representations or just the task.  Some believe that only rich syntax representations enable simple classifiers to perform well~\citep{lin2019open}. \citet{kunz_classifier_2020} overthrow these claims by demonstrating that its good performance comes from encoding local neighboring words. A study shows that classifiers rely on semantic cues fail to extract syntax~\citep{maudslay_syntactic_2021}. In contrast, other research reveals that models such as BERT encode the corresponding information in a variety of ways~\citep{mohebbi_exploring_2021, li_how_2021}. Therefore, the validity of probing syntactic information still requires further investigation. Since sentence-level syntactic information is generally distributed in each word, the prediction for them is simpler with probing classifier without dependency tree retrieval. Local syntax and semantics are usually studied together as they investigate the same objects such as neurons, layers, and contextual representation. The differences are mainly due to their training objectives and training data~\citep{tenney_bert_2019}.

The ability to learn semantic knowledge is often examined on tasks like coreference resolution, named entity labeling, relation classification, question types classification and supporting facts, etc.~\citep{van_aken_how_2019}. A prominent framework called edge probing~\citep{tenney_what_2019} has been advanced to provide exhaustive syntactic and semantic probing tools. Differently, it takes both pre-trained representation and integer spans as input and transforms them into fixed-length span representations that are fed to train a probing classifier. Because of the definition of span representation, such approach becomes extremely versatile and widely applied in syntactic and semantic probing tasks. Some works simply probe referential relations by measuring the similarity between transformed representation of pronouns and preceding words within a fixed length, and assigning a higher probability to more similar ones~\citep{sorodoc_probing_2020}. Probing work involving prompts usually faces challenges with zero-shot and few-shot learning. The evaluation with these models is more complicated as prompt quality also significantly influences performance~\citep{zhang_probing_2022}. Even with carefully designed datasets and prompt design, the result still needs further examination.

On the other hand, probing classifiers for \emph{attention heads} are designed in a similar fashion where a shallow classifier is trained on top of pre-trained models to predict certain features. Apart from relating attention heads to syntax and semantics, patterns in attention heads are also studied. A representative work trains classifiers to identify patterns using self-attention maps sampled on random inputs, then prunes heads based on this to improve model efficiency~\citep{kovaleva_revealing_2019, clark_what_2019}. Instead of making predictions, some work regards attention as semantic information indicators and traces it backward through layers, accumulates it and distributes it to input tokens to denote semantic information~\citep{wu_structured_2020}. 
But the question is whether traced attention equivalently represents semantic information across different heads.

Although high probing performance is often attributed to the quality and interpretability of representations~\citep{belinkov_probing_2022}, this assumption remains largely unproven and difficult to validate. Before we can comprehensively address such challenges, adding constraints like selectivity~\citep{hewitt_designing_2019}, which measures how selectively a probe targets the linguistic property of interest compared to an unrelated control task, may help mitigate potential probe biases in the interim.

\paragraph{Parameter-Free Probing}
There is another branch of data-centric probing techniques that does not require probing classifiers. Instead, they design datasets tailored to specific linguistic properties like grammar~\citep{marvin_targeted_2018}. The encoding model's performance illustrates its capability in capturing those properties. For language models, the measurement is based on whether the probability of positive examples is higher than that of negative examples.

Probing tasks can also be performed with data-driven prompt search, where certain knowledge are examined via language model's text generation or completion abilities~\citep{petroni_language_2019, apidianaki_all_2021, li_probing_2022}. For instance, \citet{ravichander_systematicity_2020} proves that BERT encodes hypernymy information by completing cloze tasks i.e. filling blanks in incomplete sentences in zero-shot settings. And the result demonstrates that BERT performs well in providing the right answer in the top 5 for all samples. However, as argued by~\citep{zhong_factual_2021}, training datasets include regularities that prompting methods can exploit to make predictions. The real factual knowledge captured by language models becomes obscure.

\subsubsection{Neuron Activation Explanation}\label{sec:neuron_ac}
Instead of examining the whole vector space, neuron analysis looks into individual dimensions, i.e. neurons in representations, that are crucial for model performance or associated with specific linguistic properties. 

One simple line of work follows two main steps: first, identifying important neurons in an unsupervised manner. Second, learning relations between linguistic properties and individual neurons in supervised tasks. On the assumption that different models learning similar properties usually share similar neurons, these shared neurons are ranked according to various metrics such as correlation measurements and learned weights~\citep{bau_identifying_2018, dalvi_what_2019}. Alternatively, conventional supervised classification can also be adopted to find important neurons in a given model~\citep{dalvi_what_2019}. The importance of ranked neurons is verified quantitatively via ablation experiments, e.g., masking, erasure, visualization, etc. Other probing techniques like greedy Gaussian probing have also emerged to identify important neurons~\citep{torroba_hennigen_intrinsic_2020}. However, existing methods struggle to balance accuracy and selectivity~\citep{antverg_pitfalls_2022}.

\begin{figure}[t]
    \centering
    \includegraphics[width = 1\textwidth]{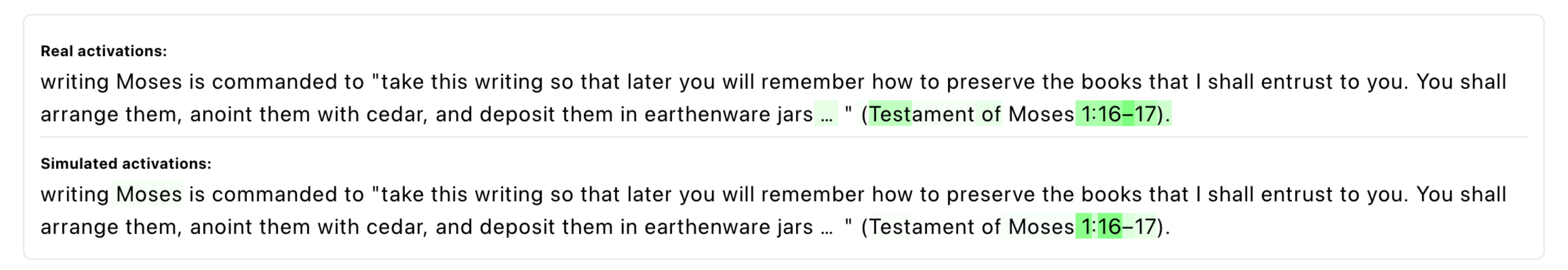}
    \caption{Activation visualization of the 131st neuron in the 5th layer of the GPT-2. The simulated explanation from GPT-4 indicates that the 131st neuron in the fifth layer of GPT-2 is activated by citations. The real activation of this neuron confirms the accuracy of the simulated explanation provided by GPT-4.}\label{fig:gpt4}
\end{figure}

Intuitively, all neurons should be examined to make explanation. However, due to the expensive computational cost and the claim that only a small subset of neurons are important for making decisions \citep{bau_identifying_2018, antverg_pitfalls_2022}
, existing approaches are always integrated with ranking algorithms. With the increasing generalization capabilities of LLMs, it becomes feasible to provide explanations for individual neurons.
A recent study by OpenAI demonstrates the use of GPT-4 to generate natural language explanations for individual neuron activation in GPT-2 XL~\citep{openAI2023}. It uses GPT-4 to summarize the patterns in text that trigger high activation values for a given GPT-2 XL neuron. For example, GPT-4 could summarize the pattern of a neuron as: \emph{references to movies, characters, and entertainment}. The quality of each neuron explanation is evaluated by testing how well it allows GPT-4 to simulate the real neuron's behavior on new text examples. Explanations are scored based on the correlation between GPT-4's simulated activation and the true activation (see Figure~\ref{fig:gpt4}). High correlation indicates an accurate explanation that captures the essence of what the neuron encodes. More than 1,000 GPT-2 XL neurons were found to have high-scoring explanations from GPT-4, which accounts for most of their behavior. This auto-generated natural language provides intuitive insight into the emergent inner computations and feature representations in GPT-2 XL.
A common limitation in explaining individual components of LLMs is the lack of ground-truth explanation annotations for individual components. Without these annotations, the evaluation of component-level explanations remains challenging.

Another recent study proposes the Summarize and Score (SASC) explanation pipeline to generate natural language explanations to explain modules from large language models~\citep{singh2023explaining}. First, SASC generates candidate explanations using a pre-trained language model to find n-grams that elicit the most positive response from the module $f$. The SASC then evaluates each candidate explanation by generating synthetic data based on the explanation and testing how $f$ responds to those data. The authors apply SASC to explain modules within BERT (bert-base-uncased), which is then compared with human-labeled explanations. The comparison indicates that SASC explanations are sometimes similar to human explanations.

\subsubsection{Concept-Based Explanation}
\label{sec:concept}
Concept-based interpretability algorithms map the inputs to a sets of concepts and measure importance score of each pre-defined concept to model predictions. By introducing abstract concepts, models can be explained in a human-understandable fashion rather than on low-level features. Information in latent space can also be transformed into comprehensible explanations. A representative framework named TCAV~\citep{kim_interpretability_2018} uses directional derivatives to quantify the contribution of defined concepts to the model predictions. It first represents the concept with a set of examples and then learns a linear classifier as ``concept activation vector''(CAV) to detect interested concept. The learned vector is used as input changes in the direction of concepts to measure prediction sensitivity with concepts i.e. importance score of concepts. Originally proposed for computer vision, TCAV has also been tailored to NLP models for sentiment classification using the IMDB sensitivity dataset~\citep{pytorchcaptum2022}. Specifically, two concepts were explored: Positive Adjectives and Neutral. The Positive Adjectives concept refers to a group of adjectives that express positive feelings. The Neutral concept spans broader domains and is distinct from Positive Adjectives. For sentences with negative sentiment, the TCAV scores indicate the Positive Adjectives score is relatively low compared to Neutral, which is consistent with human understanding. However, TCAV requires additional data to describe concepts and performance of the concept classifier can hardly be guaranteed. An alternative way of selecting concepts is to identify those learned by neurons through probing tasks with annotated datasets~\citep{mu_compositional_2021}. The study shows that neurons produce explanations not only based on individual concepts but also compositions of logical forms. The more neurons are interpretable, the more accurate the model is. A common pitfall of concept-based explanations is how to define useful concepts. Besides, it is always constrained by available descriptive datasets.

\subsubsection{Mechanistic Interpretability}

{Mechanistic interpretability understands language models by investigating individual neurons and especially their connections in terms of \emph{circuits}~\citep{anthropic2023decomposing,bricken2023towards}. Due to the motivation to regard parts of neural models as functional components, we discuss this line of work separately.}

{Circuits was originally proposed to explain vision models that are intuitive to comprehend, where detectors for complex objects can be built out of simple building blocks such as line detectors, curve detector, and so on. One stream of work studies the hidden representation of neural networks. These representations can be visualized with features. They believe that complex feature detectors can be implemented from earlier and easier feature detectors. Besides, different features can be spread across many polysemantic neurons also known as superposition~\citep{distillZoomIntroduction}. Another line of work studies weights that connect neurons aiming to find meaningful algorithms that implement simple logic. They approach subgraphs of networks with circuits denoting linear combination of features as well as logical operations, which is crucial to establish a casual relationship for predictions. Built on top of neuron-level explanation in circuits, larger-scale functional components have also been explored. Three phenomena have been identified: 1) branch specialization, 2) weight banding, 3) equivariance. Branch specialization describes the feature organization between branches, where a given type of feature was observed to group into a branch. This phenomenon exists at different levels of layers, and the same branch specialization might be robust across different architectures and tasks~\citep{distillBranchSpecialization}. The weight banding usually appears in the final layer of vision models with global average pooling~\citep{distillWeightBanding}. Equivariance captures symmetries in neural networks where many neurons are transformed from the basic version~\citep{distillNaturallyOccurring}.}

{When it comes to transformers, circuits usually work and are interpreted in a different way because of their architectures. One-layer and two-layer attention-only models have been investigated recently. For one-layer attention-only models, bigram and skip-trigram tables can be accessed from weights. However, two-layer attention-only transformers demonstrate ``induction head'' by composing attention heads from different layers~\citep{transformercircuitsMathematicalFramework}.  The induction head consists of two attention heads. The first attention head is responsible for copying information from the previous token into the next token, while the second one is used to deduct the following token with information from the first attention head. Such mechanism is believed to be the source of in-context learning, which has been demonstrated with multiple less conclusive evidences. For example, the phase change observed on cooccurrence of in-context learning and induction heads, and corresponding in-context learning shifts after perturbing or knocking out induction heads. However, due to the complex components of state-of-art language models, such as multiple layers and multilayer perceptrons, it remains to be seen whether the theory of the ``induction heads'' in these models still holds~\citep{transformercircuitsIncontextLearning}. Alternatively, some work focuses on feedforward layers that contain most of the information. Each key in transformers is taken as a memory of textual patterns in the training examples. The values induce output distribution based on keys~\citep{geva2020transformer}. By tracing the casual effects of hidden state activation within GPT and altering model weights that are decisive at model predictions, a range of middle layers are identified as relevant with facts~\citep{meng2022locating}. Another study transfers the feedforward layer as a sub-update vector which is interpreted as a small set of human-interpretable concepts~\citep{geva2022transformer}.}

{However, unlike digital circuits that have deterministic functions in each part, large-scale neural networks are more elastic and versatile in composition such as robust to remove entire layers~\citep{veit2016residual,mcgrath2023hydra}. In addition, most existing hypotheses have not been examined on large language models. Recently, \citet{lieberum2023does} explore scalability of circuit analysis in the 70B Chinchilla model. The result demonstrates activation patching~\citep{meng2022locating}, attention pattern visualization~\citep{transformercircuitsMathematicalFramework} and logit attribution can be well adapted rather than correct letter heads that moving information from the correct content tokens to the final token~\citep{lieberum2023does}. Therefore, the circuit-based explanation still requires further investigation on LLMs.}

\subsection{Making Use of Explanations}
\label{sec:using-expl-impr}
In the previous subsections, we introduced methods to generate explanations for LLMs. In this subsection, we discuss how explainability can be used as a tool to debug and improve models.

\subsubsection{Debugging Models}
\label{sec:xai-debug-models}
Post-hoc explainability methods can be used to analyze model feature importance patterns to identify biases or limitations in its behavior~\citep{du2022shortcut}. For example, if the model consistently attends to certain tokens in the input sequence regardless of the context, this may indicate that the model relies on heuristics or biases rather than truly understanding the meaning of the input sequence.
Recent work has used Integrated Gradients to debug trained language models in natural language understanding tasks, finding that they use shortcuts rather than complex reasoning for prediction~\citep{du2021towards}. Specifically, these models favor features from the head of long-tailed distributions, picking up these shortcut cues early in training. This shortcut learning harms model robustness and generalization to out-of-distribution samples.
Integrated Gradient explanations have also been used to examine the adversarial robustness of language models~\citep{chen2022adversarial}. The explanations reveal that models robust to adversarial examples rely on similar features, while non-robust models rely on different key features. These insights have motivated the development of more robust adversarial training methods.

\subsubsection{Improving Models}
\label{sec:rati-impr-model}
Regularization techniques can be used to improve the performance and reliability of model explanations. Specifically, explanation regularization (ER) methods aim to improve LLM generalization by aligning the model's machine rationales (which tokens it focuses on) with human rationales~\citep{joshi2022er}.
For example, a framework called AMPLIFY is proposed that generates automated rationales using post-hoc explanation methods~\citep{ma2023post}. 
{These automated rationales are feed as part of prompts to LLM for prediction.}
Experiments show that AMPLIFY improves the accuracy of LLMs by 10-25\% for various tasks, even when human rationale is lacking.
Another study proposes ER-TEST~\citep{joshi2022er}, a framework that evaluates the out-of-distribution (OOD) generalization of ER models along three dimensions: unseen dataset tests, contrast set tests, and functional tests. This provides a more comprehensive evaluation than just in-distribution performance. They consider three types of explainability methods, including Input*Gradient, attention-based rationale~\citep{stacey2022supervising}, and learned rationale~\citep{chan2022unirex}. Across sentiment analysis and natural language inference tasks/datasets, ER-TEST shows that ER has little impact on in-distribution performance but yields large OOD gains.
An end-to-end framework called XMD was proposed for explanation-based debugging and improvement~\citep{lee2022xmd}. XMD allows users to provide flexible feedback on task- or instance-level explanations via an intuitive interface. It then updates the model in real time by regularizing it to align explanations with user feedback. Using XMD has been shown to improve models' OOD performance on text classification by up to 18\%.

\section{Explanation for Prompting Paradigm}\label{sec:prompting-paradigm}

{As language models scale up, prompting-based models exhibit emergent abilities that require new perspectives to elucidate their underlying mechanisms. However, the aggressive surge in model scale renders traditional explanation methods unsuitable. The challenges of applying certain explanability techniques targeting traditional fine-tuning paradigms to prompting-based paradigms can be summarized from multiple facets. For example, prompting-based models rely on reasoning abilities~\citep{wei2023larger}, which makes localized or example-specific explanations much less meaningful. Additionally, computationally demanding explanation techniques quickly become infeasible at the scale of hundreds of billions of parameters or more. Further, the intricate inner workings and reasoning processes of prompting-based models are too complex to be effectively captured by simplified surrogate models.}

{In light of these challenges, new explanation techniques tailored to this prompting paradigm are emerging.
For example, chain of thought (CoT) explanations may provide a more suitable approach for understanding and explaining the behaviors of large language models based on prompting. Besides, methods that focus on identifying influential examples that contribute to predictions are gaining importance. Identifying these pivotal data points may significantly enhance our understanding of dataset composition. Global explanation techniques for traditional fine-tuning paradigms are widely employed to prompting-based LLMs as well. Particularly these techniques that capable of delivering high-level explanations such as concept-based explanation and module-based explanation. 
}

In this section, we first introduce techniques to explain models belonging to the prompting paradigm, including 1) explaining base models such as LLaMA-2 (Section~\ref{sec:base-model}), 2) explaining assistant models such as LLaMA-2-Chat (Section~\ref{sec:assistant-model}), and 3) how to harness the reasoning and explaining ability of LLMs to improve the predictive performance of language models and enable beneficial applications (Section~\ref{sec:using-explanation-prompting}).

\subsection{Base Model Explanation}
\label{sec:base-model}
As the scale of language models increases, they exhibit new abilities like few-shot learning, i.e., the ability to learn concepts from just a few examples. They also demonstrate a chain-of-thought(CoT) prompting ability, which allows feeding a sequence of prompts to the model to steer its generation in a particular direction and have it explain its reasoning~\citep{wei2022chain}. Given these emerging properties, the explainability research has three main goals: 1) 
understanding how these large language models can grasp new tasks so quickly from limited examples, which helps end-users interpret the model's reasoning, 2) explaining CoT prompting, and 3) and representation engineering.

\subsubsection{Explaining In-context Learning}\label{sec:how-context-learning}
Explainable AI techniques have been used to elucidate how prompting works in LLMs. Specifically,
we discuss techniques that shed light on how in-context learning (ICL) influences model behavior.

One study uses the SST-2 sentiment analysis benchmark as a baseline task to explain the in-context learning paradigm~\citep{li2023understanding}. It investigates how ICL works in LLMs by analyzing contrastive demonstrations and saliency maps. The authors build contrastive demonstrations by flipping labels, perturbing input text, and adding complementary explanations. For a sentiment analysis task, they find that flipping labels is more likely to reduce salience for smaller models (e.g., GPT-2), while having an opposite impact on large models (e.g., InstructGPT).
The impact of different demonstration types appears to vary depending on the model scale and task type. Further analysis is required across a range of models, tasks, and datasets.
Another study investigates whether ICL in large language models is 
enabled by semantic priors from their pre-training or if it learns input label mappings from the provided examples~\citep{wei2023larger}. Experimental results indicate that large models can override semantic priors and learn contradictory input-label mappings, while small models rely more heavily on priors. Experiments with flipped labels in ICL exemplars show that large models can learn to flip predictions, while small models cannot. These results indicate that LLMs have greater ability to learn arbitrary input-label mappings, a form of symbolic reasoning unconstrained by semantic priors, which challenges the view that ICL is solely driven by leveraging priors.

\subsubsection{Explaining CoT Prompting}
One study investigates how chain-of-thought (CoT) prompting affects the behavior of LLMs by analyzing the saliency scores of the input tokens~\citep{wu2023analyzing}. Saliency scores indicate how influential each input token is on the model's output. The scores are computed using gradient-based feature attribution methods. The goal is to understand whether CoT prompting changes saliency scores compared to standard prompting, offering insights into why CoT improves performance. The analysis of saliency scores suggests that CoT prompting makes models consider question tokens in a more stable way. This more stable consideration of input may induce the generation of more consistently accurate answers compared to standard prompting. Other work has focused on perturbing CoT demonstrations in few-shot prompts, e.g., by adding errors, to determine which aspects are important for generating high-performing explanations~\citep{madaan2022text,wang2022towards}. Counterfactual prompts have been proposed to perturb key components of a prompt: symbols, patterns, and text~\citep{madaan2022text}. Experimental analysis indicates the intermediate reasoning steps act more as a beacon for the model to replicate symbols into factual answers, rather than facilitate learning to solve the task.

\subsubsection{Representation Engineering}
{Unlike the aforementioned two lines of research that explain LLMs from the prompt engineering perspective, this family of research explains LLMs from the representation engineering perspective.
Representation engineering explains models from a top-down perspective and regards representation and its transformation as the primary element of analysis. Such approaches focus on structure and the characteristic of the representation space to capture emergent representations and high-level cognitive phenomena. \citet{zou2023representation} implement representation engineering in two parts: 1) representation reading, 2) representation control. Representation reading identifies representations for high-level concepts and functions within a network. Inspired by neuroimaging methodologies, linear artificial tomograph scan is adopted. To elicit concepts and functions well, prompt templates that include stimulus or instructions are designed separately. For concepts, neural activity can be collected either from representation of most representative tokes or from the last token. For functions, neural activity can be collected from the response after a certain token. Then, linear probes are introduced to predict concepts and functions with neural activity. Representation control aims to manipulate the inner representation of concepts and functions based on understanding from representation reading to meet safety requirements. Directly adding reading vectors can induce honest model output and subtracting reading vectors can induce models to lie, which demonstrates great potential in improving models. Similarly, studying the representation structures on a high-quality dataset of true/false statements also reveals the linear structure of representations. The trained probes generalize well on other datasets. As in the conclusion of the aforementioned work, the truth directions can be identified and used to induce true or false output~\citep{marks2023geometry}. By analyzing the learned representation of six spatial or temporal datasets, LLMs such as LLaMA-13B are demonstrated to learn linear representations of space and time. In addition, similar patterns have been found in models of different sizes. The representations are also increasingly accurate as the model scales up. The model also has specialized neurons that activate as a function of space or time, which aligns with the founding of factual knowledge in LLMs~\citep{gurnee2023language}. In conclusion, representation engineering could be promising techniques to control model output, but further ablation studies are still required to identify its strengths and weaknesses.}

\subsection{Assistant Model Explanation}
\label{sec:assistant-model}
Due to the large-scale unsupervised pre-training and the supervised alignment fine-tuning, LLMs belonging to this paradigm have strong reasoning ability. However, their sheer scale also makes them susceptible to generating problematic outputs such as hallucinations. Explainability research aims to 1) elucidate the role of the alignment fine-tuning, 2) analyze the causes of hallucinations, and 3) uncertainty quantification.

\subsubsection{Explaining the Role of Fine-tuning}
\label{sec:how-import-alignm}
Assistant models are typically trained in two stages. First, they undergo \emph{unsupervised pre-training} on large amounts of raw text to learn general linguistic representations. This pre-training stage allows the models to acquire general language knowledge. Second, the models go through \emph{alignment fine-tuning} via supervised and reinforcement learning. This aligns the models with specific end tasks and user preferences.
Explainability research on these models focuses on determining whether their knowledge comes predominantly from the initial pre-training stage, wherein they acquire general language abilities, or from the subsequent alignment fine-tuning stage, wherein they are tailored to specific tasks and preferences. Understanding the source of the models' knowledge provides insight into how to improve and interpret their performance.

A recent study by \citet{zhou2023lima} investigated the relative importance of pre-training versus instruction fine-tuning for language models. In the experiment, the authors used only 1,000 carefully selected instructions to tune the LLaMA-65B model, without reinforcement learning, and achieved performance comparable to GPT-4.
The researchers hypothesized that alignment may be a simpler process where the model learns interaction styles and formats, while almost all knowledge of LLMs is acquired during pre-training.
The experimental findings demonstrated the power of pre-training and its relative importance over large-scale fine-tuning and reinforcement learning approaches. 
Complex fine-tuning and reinforcement learning techniques may be less crucial than previously believed.
On the other hand, this study also indicates that data quality is more important compared to data quantity during instruction fine-tuning.
{Furthermore,~\citet{wu2023language} looked into the role of instruction fine-tuning by examining instruction following and concept-level knowledge evolution. The result shows that instruction fine-tuned models can better distinguish instruction and context, and follow users' instructions well. Besides, they can focus more on middle and tail of input prompts than pre-trained models. And fine-tuned models adjust concepts toward downstream user-oriented tasks explicitly but the linguistic distributions remain the same. Contradict to conventional belief that higher layers capture more semantic knowledge, the proportion of captured semantic knowledge initially grows then drops aggressively in fine-tuned models. In the view of self-attention heads activation, it is found that instruction fine-tuning adapts pre-trained models recognizing instruction verbs by making more neurons in the lower-level layer encode word-word patterns~\citep{wu2023language}.
}

Another recent study~\citep{gudibande2023false} showed that imitation can successfully improve the style, persona and ability of the language model to follow instructions, but does not improve language models on more complex dimensions such as factuality, coding, and problem solving. Imitation is another commonly used technique for training an assistant model, where a foundation model like GPT-2 or LLaMA is fine-tuned on outputs generated by a more advanced system, such as a proprietary model like ChatGPT.
Furthermore, the technical report of LLaMA-2~\citep{touvron2023llama2} suggests that the fine-tuning stage mainly helps increase the helpfulness and safety of language models, where helpfulness describes how well LLaMA-2-Chat responses satisfy user requests and contain intended information, and safety refers to avoiding unsafe responses like toxic content.

Taken together, these studies emphasize the significant role of foundation models, highlighting the importance of pre-training. The findings suggest that assistant models' knowledge is mainly captured during the pre-training stage. Subsequent instruction fine-tuning then helps activate this knowledge towards useful outputs for end users. Furthermore, reinforcement learning can further align the model with human values.

\subsubsection{Explaining Hallucination}\label{sec:underst-hall}
{The rapid development of LLMs has raised concerns about their trustworthiness, as they have the potential to exhibit undesirable behaviors such as generating hallucination, a phenomenon in which models generate output that is irrelevant and nonsensical in a natural manner~\citep{zhang_sirens_2023,huang2023look}. There emerges increasing interest from the community in understanding how hallucination is produced and how to reduce hallucination generation.} 

{Recent analysis research indicates that the hallucination phenomenon stems from various problems within datasets~\citep{dziri_origin_2022}, which can be categorized into two main classes: 1) a lack of relevant data, 2) repeated data. For example, long-tail knowledge is prevalent in training data and LLMs easily fall short in learning such knowledge~\citep{kandpal_large_2023}. 
On the other hand, deduplicating data is challenging to be done perfectly. Duplicate data within the training dataset can noticeably impair the model's performance. \citet{hernandez_scaling_2022} find that the performance of an 800M parameter model can degrade to that of a 400M parameter model by only repeating 10\% of the training data. When examining the model's performance in terms of scaling laws, a certain range of repetition frequency in the middle could have a detrimental impact. This range is hypothesized to lead the model to memorize the data and consequently consume a large portion of its capacity.}

{Moreover, recent studies find that hallucination also arises from certain limitations inherent to models. \citet{mckenna_sources_2023} demonstrate that LLMs still rely on memorization at the sentence level and statistical patterns at the corpora level instead of robust reasoning. This is evidenced by their analysis of various LLM families' performance on natural language inference tasks. Further, \citet{wu_plms_2023} reveal that LLMs are imperfect in both memorization and reasoning regarding ontological knowledge. \citet{berglund_reversal_2023} points out that LLMs usually suffer from logical deduction due to reversal curse. LLMs tend to be overconfident in their output and struggle to identify the factual knowledge boundary precisely~\citep{ren_investigating_2023}. Furthermore, LLM favors co-occurrence words over factual answers, a phenomenon often referred to as shortcuts or spurious correlations~\citep{kang_impact_2023}. Similarly, another undesired behavior sycophancy also exists in LLMs, which refers to models could generate answers aligning users' view rather than facts. The worst thing is that model scaling and instruction tuning could increase such behavior~\citep{wei_simple_2023}.
}

{There are several ways to address the hallucination problem. Firstly, scaling is always a good step to take. The performance of PaLM with 540 billion parameters steeply increased on a variety of tasks. Even it also suffers from learning long-tail knowledge, but its memorization abilities are shown to be better than small models~\citep{chowdhery_palm_2022}. In text summarization tasks, \citet{ladhak_when_2023} shows that using more extractive fine-tuning datasets and adapter-fine-tuning that fine-tunes part of parameters usually generates less hallucinations but will not change the distribution of hallucination. Consequently, mitigation can be implemented by either data side such as improve fine-tuning datasets and add synthetic-data intervention~\citep{wei_simple_2023}, or on the model side, such as different optimization approaches.}

\subsubsection{Uncertainty Quantification}
There is also growing interest within the research community in quantifying the uncertainty of LLM predictions, to better understand the reliability and limitations of these powerful models. 

Most existing literature on uncertainty quantification focuses on logits, which is however less suitable for LLMs, especially closed-source ones. This necessitates non-logit-based approaches for eliciting uncertainty in LLMs, known as confidence elicitation~\citep{xiong2023can}. There are several representative methods for uncertainty estimation of LLMs. First, consistency-based uncertainty estimation involves generating multiple responses to a question and using the consistency between these responses to estimate the model's confidence~\citep{xiong2023can}. Specifically, it introduces randomness into the answer generation process (self-consistency) or adds misleading hints to the prompt (induced-consistency) to produce varied responses. The more consistent the multiple responses are, the higher the estimated confidence in the answer is. Second, LLMs can deliver their confidence verbally by providing direct and specific responses to indicate high confidence in their predictions, and giving indirect, vague, or ambiguous responses to convey lower confidence. LLMs can explicitly state a percentage to quantify their confidence level. For example, ``I am only 20\% confident in this answer'' clearly communicates low confidence~\citep{xiong2023can}. Third, the uncertainty can be aggregated from token-level uncertainty~\citep{duan2023shifting}. LLMs generate text by predicting each token, which can be framed as a classification task. Token-level uncertainty methods calculate a confidence score for each predicted token based on its probability distribution. Then the overall uncertainty can be estimated based on the aggregation of token-level uncertainties.

\subsection{Making Use of Explanations}\label{sec:using-explanation-prompting}
In this section, we discuss techniques to harness the explanatory abilities of prompting-based LLMs to improve the predictive performance of language models and enable beneficial applications.

\subsubsection{Improving LLMs}
This line of research investigates whether LLMs can benefit from explanations when learning new tasks from limited examples. Specifically, it investigates whether providing explanations for the answers to few-shot tasks can improve the model's performance on these tasks~\citep{lampinen2022can}. Two forms of explanations are provided: \textit{pre-answer explanations} and \textit{post-answer explanations}. 
\citet{wei2022chain} proposes a method called chain-of-thought prompting, which provides intermediate reasoning steps as explanations in prompts before the answers. This has helped language models achieve state-of-the-art results in arithmetic, symbolic, and common-sense reasoning tasks.
Another recent study provides explanations after the answer in the prompts~\citep{lampinen2022can}. Experimental analysis indicates that providing explanations can improve the few-shot learning performance of large language models, but the benefits depend on model scale and explanation quality. Additionally, customizing explanations specifically for the task using the validation set further increases their benefits~\citep{lampinen2022can}. 

Another recent study proposes explanation tuning, an approach that trains smaller language models using detailed step-by-step explanations from more advanced models as a form of supervision~\citep{mukherjee2023orca}. Section~\ref{sec:how-import-alignm} indicates that imitation tuning mainly allows smaller models to learn the style of the larger models rather than the reasoning process. To address this limitation, this work proposes leveraging richer signals beyond just input-output pairs to teach smaller models to mimic the reasoning process of large foundation models like GPT-4. Specifically, the authors collect training data consisting of prompts and detailed explanatory responses from GPT-4. To allow GPT-4 to generate explanations, system instructions such as ``\texttt{You are a helpful assistant, who always provides explanation. Think like you are answering to a five-year-old.}'' are utilized. Experimental results indicate that models trained with explanation tuning outperform models trained using conventional instruction tuning in complex zero-shot reasoning benchmarks like BigBench Hard.

The insights captured from the explanations can also be utilized to compress the instructions~\citep{yin2023did}. The authors use ablation analysis to study the contribution of different categories of content in task definitions. The insights from the ablation analysis can then be utilized to compress the task instruction. Take classification task as an example, the analysis indicates that the most important components within the task instruction are label-relevant information. Removing other contents will only slightly impact the classification performance, and the authors find that model performance only drops substantially when removing output label information. Additionally, they propose an algorithm to automatically compress definitions by removing unnecessary tokens, finding 60\% can be removed while maintaining or improving performance for T5-XL model on a holdout dataset.

{Moreover, some studies have also delved into the effectiveness of the explanations generated by LLMs in enhancing few-shot in-context learning. For multi-step symbolic reasoning tasks, involving code execution and arithmetic operations, \citet{nye2021show} found that incorporating intermediate computation steps can significantly boost the model's ability. On the flip side, when it comes to textual reasoning tasks including question answering and natural language inference, only text-davinci-002 was observed an increase in accuracy. The other four models, including OPT, GPT-3(davinci), InstructGPT(text-davinci-001), and text-davinci-002, did not show a clear improvement and even performed worse. The explanations generated by LLMs are assessed in two dimensions: factuality and consistency. The result reveals that LLMs can generate unrealistic explanations but still align with predictions, which in turn leads to incorrect predictions~\citep{ye2022unreliability}. Building on top of the finding, an explanation optimization framework has been proposed to select explanations that lead to high performance~\citep{ye2023explanation}. Therefore, improving the accuracy of model predictions requires the generation of reliable explanations by LLMs, which remains a great challenge at this time.}

\subsubsection{Downstream Applications}
Explainability can also be applied to real-world problems such as education, finance, and healthcare. For example, explainable zero-shot medical diagnosis is an interesting use case.
One recent study proposes a framework for explainable zero-shot medical image classification utilizing vision-language models like CLIP along with LLMs like ChatGPT~\citep{liu2023chatgpt}.
The key idea is to leverage ChatGPT to automatically generate detailed textual descriptions of disease symptoms and visual features beyond just the disease name. This additional textual information helps to provide more accurate and explainable diagnoses from CLIP~\citep{radford2021learning}. To handle potential inaccuracies from ChatGPT on medical topics, the authors design prompts to obtain high-quality textual descriptions of visually identifiable symptoms for each disease class. Extensive experiments on multiple medical image datasets demonstrate the effectiveness and explainability of this training-free diagnostic pipeline.

\section{Explanation Evaluation}
\label{sec:eval-expl}
In previous sections, we introduced different explanation techniques and their usages, but evaluating how faithfully they reflect a model's reasoning process remains a challenge. We roughly group the evaluations into two families: evaluation of local explanation for traditional fine-tuning paradigm (Section~\ref{sec:eva_loc_ex}) and evaluation of natural language CoT explanations for prompting paradigm (Section~\ref{sec:eva_cot_ex}). Two key dimensions of evaluations are plausibility to humans and faithfulness in capturing LLMs' internal logic. 

Technically, evaluating explanation involves human or automated model approaches. Human evaluations assess plausibility through similarity between model rationales and human rationales or subjective judgments. However, these methods usually neglect faithfulness. Subjective judgments may also not align with model reasoning, making such an evaluation unreliable. As argued by \citet{jacovi_towards_2020}, faithful evaluation should have a clear goal and avoid human involvement. Automatic evaluations test importance by perturbing model rationales, avoiding human biases. Therefore, developing rigorous automatic metrics is critical for fair faithfulness evaluation, which will be covered under the faithfulness evaluation dimension.

\subsection{Explanation Evaluations in Traditional Fine-tuning Paradigms}\label{sec:eva_loc_ex}
We introduce the evaluation of the local explanation from two aspects: plausibility and faithfulness. Both parts will mainly cover universal properties and metrics that can be applied to compare various explanation approaches. We focus on quantitative evaluation properties and metrics, which are usually more reliable than qualitative evaluations.

\paragraph{Evaluating plausibility}
The plausibility of local explanation is typically measured at the input text or token level. Plausibility evaluation can be categorized into five dimensions: grammar, semantics, knowledge, reasoning, and computation~\citep{shen_interpretability_2022}. These dimensions describe the relationship between the masked input and human-annotated rationales. Different evaluation dimensions require different kinds of datasets. For example, a sentence ``\texttt{The country [MASK] was established on July 4, 1776.}'' has the human-annotated rationale ``\texttt{established on July 4, 1776}'' and the answer to the mask should be ``\texttt{the United States}'' deriving from fact/knowledge. Although rationales might be in different granularity levels such as token or snippet and dimensions, evaluation procedures are the same except for diversified metrics. Human-annotated rationales are generally from benchmark datasets, which should meet several criteria: 1) sufficiency meaning rationales are enough for people to make correct prediction; 2) compactness requiring that if any part in the rationales is removed, the prediction will change~\citep{mathew2021hatexplain}. The explanation models are then responsible for predicting important tokens and generating rationales with these tokens. The above two kinds of rationales will be measured with various metrics. Popular metrics can be classified into two classes according to their scope of measurement. Metrics measuring two token-level rationales include Intersection-Over-Union (IOU), precision and recall. Metrics that measure overall plausibility include the F1 score for discrete cases and the area under the precision recall curve (AUPRC) for continuous or soft token selection cases~\citep{deyoung_eraser_2020}. 

\paragraph{Evaluating Faithfulness} 
Evaluation principles and metrics provide a unified way to measure faithfulness quantitatively. Since they are often defined for specific explanation techniques, we will cover only some common yet universal principles from the model perspective and metrics from the data perspective.

There are several model-level principles to which explanation methods should adhere in order to be faithful, which include {implementation invariance}, {input invariance}, {input sensitivity}, {completeness}, {polarity consistency}, {prediction consistency} and {sufficiency}. 
Implementation invariance also known as model sensitivity means that the attribution scores should remain the same regardless of differences in the model architectures, as long as the networks are functionally equal~\citep{sundararajan2017axiomatic}. Even gradient-based approaches usually meet this metric well; the assumption may not be grounded.  
Input invariance requires attribution methods to reflect sensitivity of prediction models w.r.t. effective input changes. For example, attribution scores should remain the same over constant shift of the input~\citep{kindermans_reliability_2017}. Input sensitivity defines attribution scores should be non-zero for features that solely explain prediction differences~\citep{sundararajan2017axiomatic}. 
Completeness combines sensitivity and implementation invariance with path integrals from calculus~\citep{sundararajan2017axiomatic}, which only apply to differentiable approaches.
Polarity consistency points out that some high-ranking features could impose suppression effects on final predictions, which negatively impacts explanations and should be avoided, but mostly not~\citep{liu_rethinking_2022}.
Prediction consistency confines that instances with same explanations should have the same prediction. And sufficiency requires that data with same attributions should have same related labels even with different explanations~\citep{dasgupta_framework_2022}.
In this class of approaches, researchers aim at preventing certain types of contradictory explanations by formulating axioms and properties. However, each metric can address only one particular facet of faithfulness problems. It is extremely difficult to provide all-in-one solutions within a single framework. Additionally, these approaches focus solely on avoiding inconsistent behaviors of explanation models by designing properties for explanation methods. The overall performance of models is measured with the following metrics.

A prominent line of model-agnostic work measures faithfulness by quantitatively verifying the relationship between prediction and model rationales. Some common metrics calculated on the test set are as follows: 
\begin{itemize}[leftmargin=*]\setlength\itemsep{-0.3em}
    \item{{{Comprehensiveness (COMP)}}}: the change in probability for the original predicted class before and after top-ranked important tokens removed, which means how influential the rationale is. It is formulated as $\text{comprehensiveness}=m\left(x_i\right)_j-m\left(x_i \backslash r_i\right)_j$. A higher score indicates the importance of rationales/tokens~\citep{deyoung_eraser_2020}.
    \item{{{Sufficiency (SUFF)}}}: the degree to which the parts within the extracted rationales can allow the model to make a prediction, which is defined as $\text{sufficiency} =m\left(x_i\right)_j-m\left(r_i\right)_j$~\citep{deyoung_eraser_2020}.
    \item{{{Decision Flip - Fraction Of Tokens (DFFOT)}}}: the average fraction of tokens removed to trigger a decision flip~\citep{chrysostomou_improving_2021}.
    \item{{{Decision Flip - Most Informative Token (DFMIT)}}}: the rate of decision flips caused by removing the most influential token~\citep{chrysostomou_improving_2021}.
\end{itemize}
In ERASER~\citep{deyoung_eraser_2020}, related tokens are classified into groups ranked by importance scores so that tokens can be masked in a ranked order and gradually observe output changes. The correlation between output changes and the importance of masked tokens denotes models' ability in correctly attributing feature importance. As claimed by TaSc~\citep{chrysostomou_improving_2021}, higher DFMIT and lower DFFOT are preferred, as important tokens are precisely identified and models are more faithful. 
In contrast, some work measures faithfulness through weaknesses in explanations such as shortcut learning and polarity of feature importance. \citet{bastings_will_2022} quantifies the faithfulness by how well the model identifies learned shortcuts. In this case, metrics like \textit{precision@k} (the percentage of shortcuts in top-$k$ tokens) and \textit{mean rank} (the average depth searched in salience ranking) signifies how well the top features represent all ground truth shortcuts. Likewise, higher precison@k and lower mean rank indicate good faithfulness of the models. \citet{liu_rethinking_2022} examine faithfulness by performing the violation test to make sure the model correctly reflects feature importance and feature polarity.

There are two key questions that persist when evaluating explanation models, regardless of the specific metrics used: 1) how well does the model quantify important features? 2) can the model effectively and correctly extract as many influential features as possible from the top-ranked features? However, existing evaluation metrics are often inconsistent with the same explanation models. For example, the best-ranked explanation by DFFOT could be the worst with SUFF~\citep{chan2022comparative}. TaSc demonstrates that attention-based importance metrics are more robust than non-attention ones whereas regarding attention as an explanation is still debatable~\citep{jain_attention_2019}. 

Additionally, these evaluation metrics cannot be applied directly to natural language explanations, as such explanations rarely have a straightforward relationship to the inputs. \citet{atanasova_faithfulness_2023} proposes two faithfulness tests for natural language explanation models. One test is the counterfactual test, where counterfactual examples are constructed from the original example by inserting tokens that change the prediction. If words from inserted tokens are not present in the explanation, the explanation approach is deemed unfaithful. Another test is the input reconstruction test, which explores whether the explanation is sufficient to make the same prediction as the original example. The explanation for each example is transformed into a new input given the original input and the explanation itself. Unfortunately, because both tests can introduce new linguistic variants, they struggle with evaluating faithfulness fairly when new phrases are generated. Alternatively, Rev~\citep{chen2022rev} provides evaluation metrics from the perspective of information by examining whether natural language explanations support model predictions and whether new information from explanations justify model predictions.

\subsection{Evaluation of Explanations in Prompting Paradigms}\label{sec:eva_cot_ex}
Recently, LLMs such as GPT-3 and GPT-4 have shown impressive abilities to generate natural language explanations for their predictions. However, it remains unclear whether these explanations actually help humans understand the model's reasoning process and generalize to new inputs. Note that the goals and perspectives of evaluating such explanations (e.g., CoT rationales) are different from those of evaluating traditional explanations introduced in Section~\ref{sec:eva_loc_ex}~\citep{golovneva2022roscoe,prasad2023receval}. Metrics such as plausibility, faithfulness and stability also known as diversity have been developed to evaluate explanation. Similar to traditional explanations, we focus on evaluating plausibility and faithfulness.

\paragraph{Evaluating Plausibility}
One recent work studies whether explanations satisfy human expectations and proposes to evaluate the counterfactual simulatability of natural language explanations~\citep{chen2023models}. That is, whether an explanation helps humans infer how an AI model will behave on diverse counterfactual inputs. They implement two metrics: simulation generality (diversity of counterfactuals the explanation helps simulate) and simulation precision (fraction of simulated counterfactuals where human guess matches model output). They find that explanations from LLMs such as GPT-3.5 and GPT-4 have low precision, indicating that they mislead humans to form incorrect mental models. The paper reveals limitations of current methods and that optimizing human preferences like plausibility may be insufficient for improving counterfactual simulatability. 

\paragraph{Evaluating Faithfulness}
This line of studies the faithfulness of explanations, i.e., examining how well explanations reflect the actual reasons behind a model's predictions. For example, experimental analysis of one recent study indicates that the chain-of-thought explanation can be systematically unfaithful~\citep{turpin2023language}. The authors introduced bias into model inputs by reordering multiple choice options in few-shot prompts to make the answer always ``(A)". However, language models like GPT-3.5 and Claude 1.0 failed to acknowledge the influence of these biased features in their explanations. The models generated explanations that did not faithfully represent the true decision-making process. {Another work also indicates that the LLM’s stated CoT reasoning could be unfaithful on some tasks, and smaller models tend to generate more faithful explanations compared to larger and more capable models~\citep{lanham2023measuring}.} These research highlights concerns about the faithfulness of explanations from LLMs, even when they appear sensible. {To improve reasoning faithfulness over CoT, one preliminary study proposes to generate models reasoning by decomposing questions into subquestions and answering them separately~\citep{radhakrishnan2023question}. The analysis indicates that decomposition methods can approach CoT's performance while increasing faithfulness on several metrics.} More future research is needed to develop methods to make model explanations better reflect the underlying reasons for predictions.

\section{Research Challenges}
\label{sec:research-challenges}
In this section, we explore key research challenges that warrant further investigation from both the NLP and the explainable AI communities. 

\subsection{Explanation without Ground Truths}
Ground truth explanations for LLMs are usually inaccessible. For example, there are currently no benchmark datasets to evaluate the global explanation of individual components captured by LLMs.
This presents two main challenges. First, it is difficult to design explanation algorithms that accurately reflect an LLM's decision-making process. Second, the lack of ground truth makes evaluating explanation faithfulness and fidelity problematic.
It is also challenging to select a suitable explanation among various methods in the absence of ground truth guidance. Potential solutions include involving human evaluations and creating synthetic explanatory datasets.

\subsection{Sources of Emergent Abilities}
LLMs exhibit surprising new capabilities as the model scale and training data increases, even without being explicitly trained to perform these tasks. Elucidating the origins of these emergent abilities remains an open research challenge, especially for proprietary models like ChatGPT and Claude whose architectures and training data are unpublished. Even open-source LLMs like LLaMA currently have limited interpretability into the source of their emergent skills. This can be investigated from both a model and a data perspective.

\paragraph{Model Perspective}
It is crucial to further investigate the Transformer-based model to shed light on the inner workings of LLMs. Key open questions include: 1) What specific model architectures give rise to the impressive emergent abilities of LLMs? 2) What is the minimum model complexity and scale needed to achieve strong performance across diverse language tasks? Continuing to rigorously analyze and experiment with foundation models remains imperative as LLMs continue to rapidly increase in scale. Advancing knowledge in these areas will enable more controllable and reliable LLMs. This can provide hints as to whether there will be new emergent abilities in the near future.

\paragraph{Data Perspective}
In addition to the model architecture, training data is another important perspective for understanding the emergent abilities of LLMs. Some representative research questions include: 1) Which specific subsets of the massive training data are responsible for particular model predictions, and is it possible to locate these examples? 2) Are emergent abilities the result of model training or an artifact of data contamination issues~\citep{blevins2023prompting}? 3) Are training data quality or quantity more important for effective pre-training and fine-tuning of LLMs? Understanding the interplay between training data characteristics and the resulting behavior of the model will provide key insights into the source of emergent abilities in large language models. 

\subsection{Comparing Two Paradigms}
For a given task such as natural language inference (NLI), the downstream fine-tuning paradigm and prompting paradigm can demonstrate markedly different in-distribution and out-of-distribution (OOD) performance. This suggests that the two approaches rely on divergent reasoning for predictions. However, a comprehensive comparison of explanations between fine-tuning and prompting remains lacking. Further research is needed to better elucidate the explanatory differences between these paradigms. Some interesting open questions include: 1) How do fine-tuned models and prompted models differ in the rationales used for prediction on in-distribution examples? and 2) What causes the divergence in OOD robustness between fine-tuning and prompting? Can we trace this back to differences in reasoning?
Advancing this understanding will enable selecting the right paradigm for given use cases and improving robustness across paradigms.

\subsection{Shortcut Learning of LLMs}
Recent explainability research indicates that language models often take shortcuts when making predictions. For the downstream fine-tuning paradigm, studies show that language models leverage various dataset artifacts and biases for natural language inference tasks, such as lexical bias, overlap bias, position bias, and style bias~\citep{du2022shortcut}. This significantly impacts out-of-distribution generalization performance. For the prompting paradigm, a recent study analyzes how language models use longer contexts~\citep{liu2023lost}. The results show that performance was highest when relevant information was at the beginning or end of the context, and worsened when models had to access relevant information in the middle of long contexts. These analyses demonstrate that both paradigms tend to exploit shortcuts in certain scenarios, highlighting the need for more research to address this problem and improve generalization capabilities.

\subsection{Attention Redundancy}
Recent research has investigated attention redundancy using interpretability techniques in large language models for both traditional fine-tuning and prompting paradigms~\citep{bian2021attention,bansal2022rethinking}. For example, Bian et al. analyze attention redundancy across different pretraining and fine-tuning phases using BERT-base~\citep{bian2021attention}. Experimental analysis indicates that there is attention redundancy, finding that many attention heads are redundant and could be pruned with little impact on downstream task performance. Similarly, Bansal et al. investigate attention redundancy in terms of the in-context learning scenario using OPT-66B~\citep{bansal2022rethinking}. They found that there is redundancy in both attention heads and feedforward networks. Their findings suggest that many attention heads and other components are redundant. This presents opportunities to develop model compression techniques that prune redundant modules while preserving performance on downstream tasks.

\subsection{Shifting from Snapshot Explainability to Temporal Analysis}
{There is also an viewpoint that current interpretability research neglect the training dynamics. Existing research is mainly post-hoc explanation on fully trained models. The lack of developmental investigation on training process can generate biased explanation by failing in targeting emerging abilities or vestigial parts that convergence counts on, namely phase transitions. Besides, performing interventions on certain features fail to reflect interactions between features~\citep{nsaphraInterpretabilityCreationism}. Therefore, there is a trend shifting from static, snapshot explainability analysis to dynamic, temporal analysis. By examining several checkpoints during training, \citet{chen2023sudden} identified an abrupt pre-training window wherein models gain Syntactic Attention Structure (SAS), which occurs when a specialized attention head focus on a word's syntactic neighbors, and meanwhile a steep drop in training loss. They also showed that SAS is critical for acquiring grammatical abilities during learning. Inspired by such a perspective, development analysis could uncover more casual relations and training patterns in the training process that are helpful in understanding and improving model performance.}

\subsection{Safety and Ethics}
The lack of interpretability in LLMs poses significant ethical risks as they become more capable. Without explainability, it becomes challenging to analyze or constrain potential harms from issues such as misinformation, bias, and social manipulation. Explainable AI techniques are vital to audit these powerful models and ensure alignment with human values. For example, tools to trace training data attribution or visualize attention patterns can reveal embedded biases, such as gender stereotypes~\citep{li2023survey}. Additionally, probing classifiers can identify if problematic associations are encoded within the model's learned representations. Researchers, companies, and governments deploying LLMs have an ethical responsibility to prioritize explainable AI. Initiatives such as rigorous model audits, external oversight committees, and transparency regulations can help mitigate risks as LLMs become more prevalent. For example, as alignment systems continue to grow in scale, human feedback is becoming powerless at governing them, posing tremendous challenges for the safety of these systems. As claimed by \cite{lesswrongLevelsAlignment}, leveraging explainability tools as part of audit processes to supplement human feedback could be a productive approach. Advancing interpretability techniques must remain a priority alongside expanding model scale and performance to ensure the safe and ethical development of increasingly capable LLMs.

\section{Conclusions}
\label{sec:concl-future-work}
In this paper, we have presented a comprehensive overview of explainability techniques for LLMs. We summarize methods for local and global explanations based on model training paradigms. We also discuss using explanations to improve models, evaluation, and key challenges. Major future development options include developing explanation methods tailored to different LLMs, evaluating explanation faithfulness, and improving human interpretability. As LLMs continue to advance, explainability will become incredibly vital to ensure these models are transparent, fair, and beneficial. We hope that this survey provides a useful organization of this emerging research area, as well as highlights open problems for future work.

\bibliography{tmlr,group-references}
\bibliographystyle{tmlr}

\end{document}